\documentclass[10pt,twocolumn,letterpaper]{article}

\usepackage[pagenumbers]{cvpr} 

\usepackage{array}
\usepackage{colortbl}
\usepackage{multirow, makecell}
\usepackage{subcaption}
\usepackage{booktabs}
\usepackage{amsmath}
\usepackage{amssymb}
\usepackage{amsfonts}
\usepackage{enumitem}
\usepackage{dsfont}
\usepackage[dvipsnames]{xcolor}

\definecolor{cvprblue}{rgb}{0.21,0.49,0.74}
\usepackage[pagebackref,breaklinks,colorlinks,citecolor=cvprblue]{hyperref}

\def\paperID{15521} 
\def\confName{CVPR}
\def\confYear{2024}

\title{
\modelname: Text-to-Image Generation With Realistic Hand Appearances 
}

\author{
Supreeth Narasimhaswamy\thanks{Work started when Supreeth was an intern at Adobe Research}$^{*1}$, Uttaran Bhattacharya$^2$, Xiang Chen$^2$, \\ Ishita Dasgupta$^2$, Saayan Mitra$^2$, and Minh Hoai$^1$ \\ \\
$^1$Stony Brook University, USA, $^2$Adobe Research, USA
}

\begin{document}
\newcommand{\sn}[1]{\textcolor{blue}{[SN: {#1}]}}

\newcommand{\modelname}{HanDiffuser}
\newcommand{\compA}{Text-to-Hand-Params}
\newcommand{\compAshort}{T2H}
\newcommand{\compB}{Text-Guided Hand-Params-to-Image}
\newcommand{\compBshort}{T-H2I}

\def\mA{\mathcal{A}}
\def\mB{\mathcal{B}}
\def\mC{\mathcal{C}}
\def\mD{\mathcal{D}}
\def\mE{\mathcal{E}}
\def\mF{\mathcal{F}}
\def\mG{\mathcal{G}}
\def\mH{\mathcal{H}}
\def\mI{\mathcal{I}}
\def\mJ{\mathcal{J}}
\def\mK{\mathcal{K}}
\def\mL{\mathcal{L}}
\def\mM{\mathcal{M}}
\def\mN{\mathcal{N}}
\def\mO{\mathcal{O}}
\def\mP{\mathcal{P}}
\def\mQ{\mathcal{Q}}
\def\mR{\mathcal{R}}
\def\mS{\mathcal{S}}
\def\mT{\mathcal{T}}
\def\mU{\mathcal{U}}
\def\mV{\mathcal{V}}
\def\mW{\mathcal{W}}
\def\mX{\mathcal{X}}
\def\mY{\mathcal{Y}}
\def\mZ{\mathcal{Z}} 

\def\bbN{\mathbb{N}} 
\def\bbR{\mathbb{R}} 
\def\bbP{\mathbb{P}} 
\def\bbQ{\mathbb{Q}} 
\def\bbE{\mathbb{E}}

\def\1n{\mathbf{1}_n}
\def\0{\mathbf{0}}
\def\1{\mathbf{1}}

\def\A{{\bf A}}
\def\B{{\bf B}}
\def\C{{\bf C}}
\def\D{{\bf D}}
\def\E{{\bf E}}
\def\F{{\bf F}}
\def\G{{\bf G}}
\def\H{{\bf H}}
\def\I{{\bf I}}
\def\J{{\bf J}}
\def\K{{\bf K}}
\def\L{{\bf L}}
\def\M{{\bf M}}
\def\N{{\bf N}}
\def\O{{\bf O}}
\def\P{{\bf P}}
\def\Q{{\bf Q}}
\def\R{{\bf R}}
\def\S{{\bf S}}
\def\T{{\bf T}}
\def\U{{\bf U}}
\def\V{{\bf V}}
\def\W{{\bf W}}
\def\X{{\bf X}}
\def\Y{{\bf Y}}
\def\Z{{\bf Z}}

\def\a{{\bf a}}
\def\b{{\bf b}}
\def\c{{\bf c}}
\def\d{{\bf d}}
\def\e{{\bf e}}
\def\f{{\bf f}}
\def\g{{\bf g}}
\def\h{{\bf h}}
\def\i{{\bf i}}
\def\j{{\bf j}}
\def\k{{\bf k}}
\def\l{{\bf l}}
\def\m{{\bf m}}
\def\n{{\bf n}}
\def\o{{\bf o}}
\def\p{{\bf p}}
\def\q{{\bf q}}
\def\r{{\bf r}}
\def\s{{\bf s}}
\def\t{{\bf t}}
\def\u{{\bf u}}
\def\v{{\bf v}}
\def\w{{\bf w}}
\def\x{{\bf x}}
\def\y{{\bf y}}
\def\z{{\bf z}}

\def\balpha{\mbox{\boldmath{$\alpha$}}}
\def\bbeta{\mbox{\boldmath{$\beta$}}}
\def\bdelta{\mbox{\boldmath{$\delta$}}}
\def\bgamma{\mbox{\boldmath{$\gamma$}}}
\def\blambda{\mbox{\boldmath{$\lambda$}}}
\def\bsigma{\mbox{\boldmath{$\sigma$}}}
\def\btheta{\mbox{\boldmath{$\theta$}}}
\def\bomega{\mbox{\boldmath{$\omega$}}}
\def\bxi{\mbox{\boldmath{$\xi$}}}
\def\bnu{\mbox{\boldmath{$\nu$}}}                                  
\def\bphi{\mbox{\boldmath{$\phi$}}}
\def\bmu{\mbox{\boldmath{$\mu$}}}

\def\bDelta{\mbox{\boldmath{$\Delta$}}}
\def\bOmega{\mbox{\boldmath{$\Omega$}}}
\def\bPhi{\mbox{\boldmath{$\Phi$}}}
\def\bLambda{\mbox{\boldmath{$\Lambda$}}}
\def\bSigma{\mbox{\boldmath{$\Sigma$}}}
\def\bGamma{\mbox{\boldmath{$\Gamma$}}}
                                  
\newcommand{\myprob}[1]{\mathop{\mathbb{P}}_{#1}}

\newcommand{\myexp}[1]{\mathop{\mathbb{E}}_{#1}}

\newcommand{\mydelta}[1]{1_{#1}}

\newcommand{\parens}[1]{\left(#1\right)}
\newcommand{\braces}[1]{\left\{#1\right\}}
\newcommand{\bracks}[1]{\left[#1\right]}
\newcommand{\modulus}[1]{\left\vert#1\right\vert}
\newcommand{\norm}[1]{\left\Vert#1\right\Vert}
\newcommand{\angular}[1]{\langle#1\rangle}
\newcommand{\lmod}{\left|\!\left|}
\newcommand{\rmod}{\right|\!\right|}

\newcommand{\myminimum}[1]{\mathop{\textrm{minimum}}_{#1}}
\newcommand{\mymaximum}[1]{\mathop{\textrm{maximum}}_{#1}}    
\newcommand{\mymin}[1]{\mathop{\textrm{minimize}}_{#1}}
\newcommand{\mymax}[1]{\mathop{\textrm{maximize}}_{#1}}
\newcommand{\mymins}[1]{\mathop{\textrm{min.}}_{#1}}
\newcommand{\mymaxs}[1]{\mathop{\textrm{max.}}_{#1}}  
\newcommand{\myargmin}[1]{\mathop{\textrm{argmin}}_{#1}} 
\newcommand{\myargmax}[1]{\mathop{\textrm{argmax}}_{#1}} 
\newcommand{\myst}{\textrm{s.t. }}

\newcommand{\denselist}{\itemsep -1pt}
\newcommand{\sparselist}{\itemsep 1pt}





\def\changemargin#1#2{\list{}{\rightmargin#2\leftmargin#1}\item[]}
\let\endchangemargin=\endlist
                                               
\newcommand{\cm}[1]{}

\newcommand{\mhoai}[1]{{\color{magenta}\textbf{[MH: #1]}}}
\newcommand{\ub}[1]{{\color{BurntOrange}\textbf{[UB: #1]}}}

\newcommand{\mtodo}[1]{{\color{red}$\blacksquare$\textbf{[TODO: #1]}}}
\newcommand{\myheading}[1]{\vspace{1ex}\noindent \textbf{#1}}
\newcommand{\htimesw}[2]{\mbox{$#1$$\times$$#2$}}

\newcommand{\young}[1]{{\color{blue}$\blacksquare$\textbf{Alternative}: #1}}


\newif\ifshowsolution
\showsolutiontrue

\ifshowsolution  
\newcommand{\Comment}[1]{\paragraph{\bf $\bigstar $ COMMENT:} {\sf #1} \bigskip}
\newcommand{\Solution}[2]{\paragraph{\bf $\bigstar $ SOLUTION:} {\sf #2} }
\newcommand{\Mistake}[2]{\paragraph{\bf $\blacksquare$ COMMON MISTAKE #1:} {\sf #2} \bigskip}
\else
\newcommand{\Solution}[2]{\vspace{#1}}
\fi

\newcommand{\truefalse}{
\begin{enumerate}
	\item True
	\item False
\end{enumerate}
}

\newcommand{\yesno}{
\begin{enumerate}
	\item Yes
	\item No
\end{enumerate}
}

\newcommand{\Sref}[1]{Sec.~\ref{#1}}
\newcommand{\Eref}[1]{Eq.~(\ref{#1})}
\newcommand{\Fref}[1]{Fig.~\ref{#1}}
\newcommand{\Tref}[1]{Table~\ref{#1}}

\maketitle


\begin{abstract}

Text-to-image generative models can generate high-quality humans, but realism is lost when generating hands. Common artifacts include irregular hand poses, shapes, incorrect numbers of fingers, and physically implausible finger orientations. To generate images with realistic hands, we propose a novel diffusion-based architecture called \modelname~that achieves realism by injecting hand embeddings in the generative process. \modelname~consists of two components: a \compA~diffusion model to generate SMPL-Body and MANO-Hand parameters from input text prompts, and a \compB~diffusion model to synthesize images by conditioning on the prompts and hand parameters generated by the previous component. We incorporate multiple aspects of hand representation, including 3D shapes and joint-level finger positions, orientations and articulations, for robust learning and reliable performance during inference. We conduct extensive quantitative and qualitative experiments and perform user studies to demonstrate the efficacy of our method in generating images with high-quality hands. Project page: \url{https://supreethn.github.io/research/handiffuser/index.html}
\vskip -.1in

\end{abstract}

\section{Introduction}

\begin{figure*}[t]
    \centering
    \includegraphics[width=\textwidth]{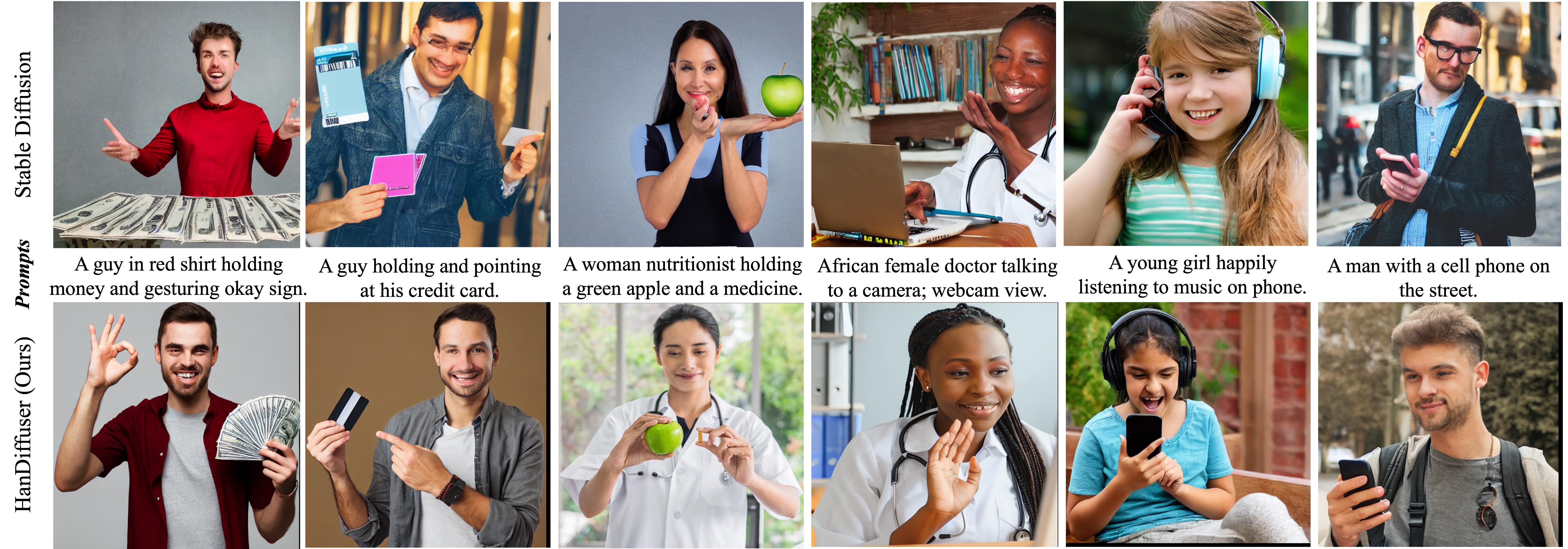}
    \caption{\textbf{Generating realistic hands.} Text-to-Image generative models, \textit{e.g.}, \cite{Rombach_2022_CVPR}, often produce various hand artifacts \textit{(top row)}. We inject hand embeddings, capturing hand shapes, poses, and articulations, in the generation process to generate realistic hands \textit{(bottom row)}.}
    \label{fig:teaser}
    \vskip -.1in

\end{figure*}

Text-to-Image (T2I) generative models have shown impressive advancement in recent years. Generative models such as Stable Diffusion~\cite{Rombach_2022_CVPR}, Imagen~\cite{Imagen_NIPS_2022}, and GLIDE~\cite{GLIDE} can generate high quality, photorealistic images. However, these methods often struggle to synthesize high-quality and realistic hands. The generated hands often have improbable hand poses, irregular hand shapes, incorrect number of fingers, and poor hand-object interactions (\Fref{fig:teaser}).

Generating images with high-quality hands is a challenging problem since hands often take up a small part of the image, but are highly articulate. They have high degrees of freedom, with a wide variety of flexibility where fingers can bend to various degrees relatively independently. Hands can also occur in various shapes, sizes, and orientations and can be occluded with other human body parts. Further, hands often interact with objects and can have a wide range of grasps depending on the object's size, shape, and affordance. Therefore, capturing such a vast range of articulations and interactions directly from text inputs remains challenging. Despite having billions of parameters and several millions of trainable images, T2I models struggle to generate realistic hands.

A central challenge in hand image generation is learning diverse hand poses and configurations at scale. Existing hand representations based on keypoint skeletons and shape formats~\cite{SMPL_2015,MANO_Hand} are useful for generative tasks in pose animation~\cite{MDM} and hand-object interactions~\cite{ARCTIC}. These representations provide a grounded understanding of plausible hand shapes and postures, especially in relation to the rest of the body and different interacting objects. However, the necessary steps to incorporate these hand representations into T2I pipelines, in terms of both learning these representations from text prompts and mapping these representations into the pixel space of images, remain open problems. These problems are exacerbated when we consider naturally constructed prompts, which often imply rather than specify hand postures and articulations (\textit{e.g.}, all prompts in \Fref{fig:teaser}). Prompt engineering~\cite{prompt_engg1,prompt_engg2}, focusing on hand descriptions, can potentially improve the generation quality. But it comes with the cost of distilling and learning appropriate prompts from large-scale data, and with the caveats of learning spurious inter-relationships between the prompt and the hands or between the hands and the rest of the images.

In this paper, we propose a learning-based model to generate images containing realistic hands in an end-to-end fashion from text prompts. Our model, called \modelname, consists of two key trainable components.
The first component, \compA~(\compAshort), generates parameters of a hand model~\cite{SMPL_2015, MANO_Hand} conditioned on the input text prompts. The second component, \compB~(\compBshort), uses the hand parameters and the input text prompts as conditions to generate images. By conditioning the image generation on accurate hand models, \modelname~can generate high-quality hands with plausible hand poses, shapes, and finger articulations.
Specifically, we consider three aspects of hand representation, each serving a unique purpose. These include the spatial locations of hand joints to capture the hand pose, the joint rotations to capture the finger orientations and articulations, and the hand vertices to capture the overall hand shape. We design a novel Text+Hand Encoder by extending the CLIP encoder~\cite{clip} to obtain joint embeddings for these three representations together with the text. We use the proposed joint embeddings to condition the image generation, allowing us to generate images by conditioning on both the hand parameters and the text.

We train the two components of \modelname~independently. We train \compAshort~using around 450K text and 3D human pairs and fine-tune \compBshort~using around 900K text and image pairs. Once trained, we use the two components end-to-end in a single inference pipeline to generate images from text prompts. We conduct extensive experiments and user studies to show the effectiveness of the \modelname~in generating images with high-quality hands.

In short, the contributions of our paper are: 

\begin{itemize}
    \item \textbf{\modelname}, a generative model to synthesize images with high-quality hands by conditioning on text and hand embeddings. It has two novel components: \compA~and \compB.

    \item \textbf{\compA}, a diffusion model to generate SMPL-Body and MANO-Hand parameters from text inputs. The generated MANO-Hands are used to further condition the image generation.
    
    \item \textbf{\compB}, a diffusion model to generate images with high-quality hands by conditioning on hand and text embeddings. We design hand embeddings to capture hand shape, pose, and finger orientations and articulations.
    
\end{itemize}
\begin{figure*}[t]
\centering
\includegraphics[width=0.98\linewidth]{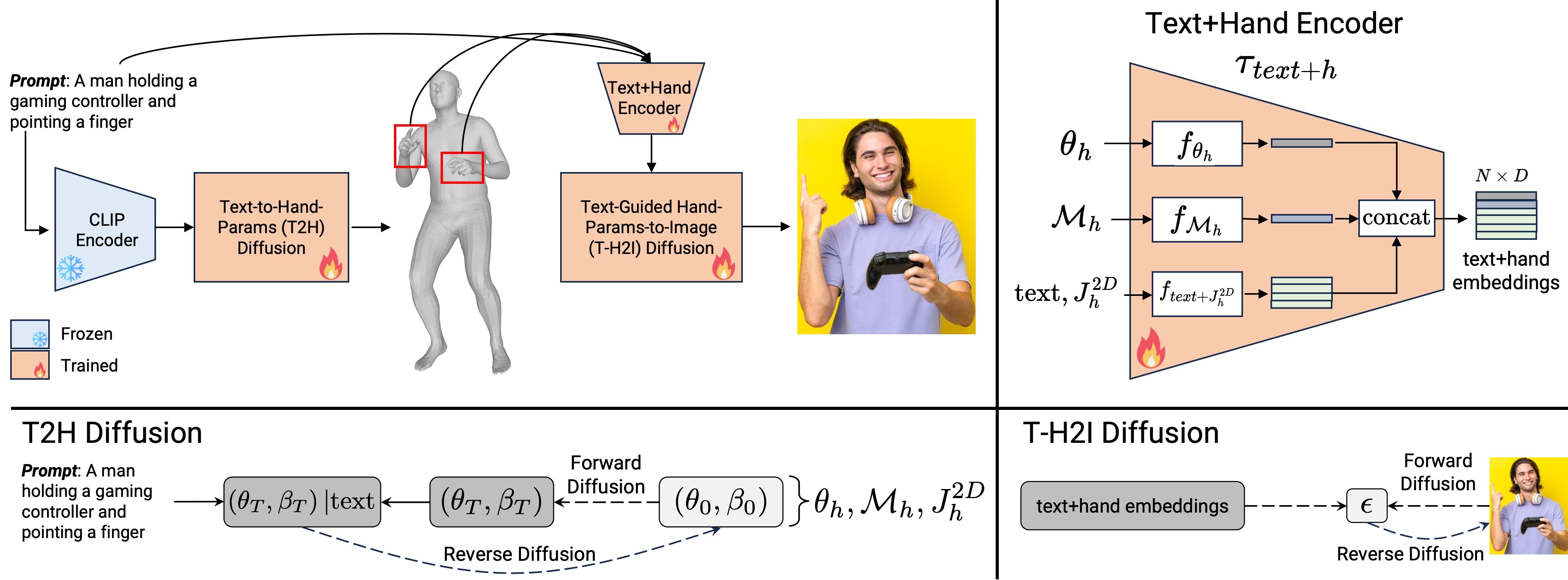}
\vskip -.1in
\caption{\textbf{\modelname~architecture.} Our architecture consists of two components. The first component, \compA~(\compAshort), takes the text as input and generates body and hand parameters. The second component, \compB~(\compBshort), uses the hand parameters from the first component and the text to generate images with high-quality hands. The Text+Hand encoder jointly encodes hand parameters and text, and captures hand pose, articulation, and shape.} 
\label{fig:architecture}
\vskip -.1in
\end{figure*}

\section{Related Work}
We briefly summarize related work on text-to-image generation, concurrent work on text-to-human generation, and commonly used hand representations.

\myheading{Text-to-Image Generation.} Text-guided image generation is a well-studied problem, with modern approaches ranging from GANs~\cite{AttnGAN,dm-gan,t2i-cl,df-gan}, autoregressive generation approaches~\cite{autoregress_attn,t2i-zs}, and VQ-VAE transformers~\cite{make_a_scene} to state-of-the-art diffusion models~\cite{image_diffusion,ddpm,diff_beats_gans,cascade_diff}. Text-to-image generation using diffusion models often bootstrap the generative pipeline with pre-trained language models, such as BERT~\cite{bert} or CLIP~\cite{clip}, to efficiently learn from the text information~\cite{t2i-cm,sdedit,t2i-vq,GLIDE,disco-diff}. More recently, Stable Diffusion~\cite{Rombach_2022_CVPR} performs diffusion in the latent image space to generate high-quality images at low computational costs. Imagen~\cite{Imagen_NIPS_2022}, by contrast, diffuses the pixels directly in a hierarchical fashion. ControlNet~\cite{controlnet} provides additional controllability in the image generation process in the form of conditioning signals ranging from sketches to pose priors. Latest software products for text-to-image generation include Midjourney~\cite{MJ}, DALL-E 3~\cite{dalle3}, and Firefly~\cite{firefly}. While the advancements in this area have been rapid and significant, generating highly articulate hands remains prone to unrealistic artifacts.

\myheading{Text-to-Human Generation.} Alongside image generation, there has also been considerable progress in human pose and motion generation from text prompts. Recent generative methods typically follow various skeletal joint formats, such as OpenPose~\cite{openpose}, or combined joint and mesh formats, such as SMPL~\cite{SMPL_2015}, to represent the human body. They train on various large-scale pose and motion datasets, including KIT~\cite{kit-mocap}, AMASS~\cite{amass}, BABEL~\cite{babel}, and HumanML3D~\cite{humanml3d}. To efficiently map text prompts to motion sequences, text-to-motion generation methods learn combined representations of language and pose using techniques ranging from recurrent neural networks~\cite{text2action,lin2018motion}, hierarchical pose embeddings~\cite{lang2pose,comp_anim}, and VQ-VAE transformers~\cite{actor,temos} to motion diffusion models~\cite{motiondiffuse,MDM,freeform_l2m,mofusion}. Some approaches even generate 3D meshes on top of pose motion synthesis to synthesize fully rendered humans~\cite{avatarclip,clip_actor}. Separate from motion synthesis, there are works on generating parametric pose models from text~\cite{delmas2022posescript,protores,delmas2023posefix}.
However, these methods focus on the human body and ignore the hand regions. As a result, they cannot generate articulate hands. There is a method~\cite{sn_handdiffusion_iccv_2023} to generate plausible hands using ControlNet but it requires a hand skeleton or mesh as an additional input.  A concurrent work \cite{lu2023handrefiner} proposes an inpainting approach to refine hands. Given a generated image, the method reconstructs 3D meshes for hands and further refines hand regions. As a result, the quality of the reconstructed hand mesh, and consequently the final refined hand, depends on the quality of the initially generated hand. On the contrary, our method first generates hand mesh parameters from the prompt and further conditions the image generation on such intermediate hand parameters. Moreover, ~\cite{lu2023handrefiner} ignores the hand-object interactions in the initial image and might not preserve hand-object occlusions and interactions when refining hands.

\myheading{Hand Representations.} Available datasets on hand configurations, gestures, and hand-object interactions offer hand representations in a variety of formats, including bounding boxes, silhouettes, depth maps~\cite{Narasimhaswamy_2019_ICCV,cslr_dataset,ntu_rgbd, bodyhands_2022, contacthands_2020, handler_2022, shilkrot_bmvc_2019, sn_hoist_cvpr_2024}, keypoints and parametric models~\cite{MANO_Hand,hagrid,affordpose,ARCTIC}. These representations are useful for multiple hand-centric tasks, including detection~\cite{yolov8}, gesture and pose recognition~\cite{Zheng_2023_CVPR}, motion generation~\cite{toch,grip}, and hand-object interactions~\cite{hoi_image,goal_hoi,imos}. Our work combines representations based on keypoint and parametric models to efficiently encode diverse hand shapes and highly articulated finger movements.

\section{\modelname}

\Fref{fig:architecture} illustrates the proposed \modelname~architecture. Given a text input, \modelname~first uses a novel \compA~diffusion model to generate the parameters of the human body and hand models. The second component is the \compB~diffusion model that generates the output image by conditioning on the hand model and the text. This section provides more detailed insights into the \compA~and \compB~models, following a brief introduction to the fundamentals of human models and stable diffusion.

\setlength{\tabcolsep}{9pt}

\subsection{Preliminaries}

\myheading{SMPL-H.} Our \compA~model generates parameters of human body and hand models from text inputs. We use SMPL~\cite{SMPL_2015} and MANO~\cite{MANO_Hand} as our body and hand model, respectively. The SMPL is a differentiable function $\mathcal{M}_b(\theta_{b}, \beta_{b})$ that takes a pose parameter $\theta_b \in \mathbb{R}^{69}$ and shape parameter $\beta_b \in \mathbb{R}^{10}$, and returns the body mesh $\mathcal{M}_b \in \mathbb{R}^{6890 \times 3}$ with 6890 vertices. Similarly, MANO is a differentiable function $\mathcal{M}_h(\theta_h, \beta_h, s)$ that takes the hand pose parameter $\theta_h \in \mathbb{R}^{48}$, hand shape parameter $\beta_h \in \mathbb{R}^{10}$, and the hand side $s \in \{\text{left}, \text{right}\}$, and returns hand mesh $\mathcal{M}_h \in \mathbb{R}^{778 \times 3}$ with 778 vertices. The 3D hand joint locations $J_h \in \mathbb{R}^{k \times 3} = \mathcal{W}_h \mathcal{M}_h$ can be regressed from vertices using a pre-trained linear regressor $\mathcal{W}_h$. The SMPL-H model combines the body, left hand, and right hand model into a single differntiable function $\mathcal{M}(\theta, \beta)$ with pose parameters $\theta {=} (\theta_b, \theta_{lh}, \theta_{rh})$ and shape parameters $\beta$. The pose parameters $\theta_b$, $\theta_{lh}$, and $\theta_{rh}$ captures the root-relative joint rotations for body, left hand, and right hand, respectively. The shape parameter $\beta$ captures the scale of the person.

\myheading{Stable Diffusion.} Our \compB~ model is built upon Stable Diffusion~\cite{Rombach_2022_CVPR}. Stable Diffusion is a latent diffusion model consisting of an auto-encoder, a U-Net for noise estimation, and a CLIP text encoder. The encoder $\mathcal{E}$ encodes an image $x$ into a latent representation $z = \mathcal{E}(x)$ that the diffusion process operates on. The decoder $\mathcal{D}$ reconstructs the image from $\hat{x} = \mathcal{D}(z)$ from the latent $z$. The U-Net is conditioned on the denoising step~$t$ and the text $\tau_{text}(text)$, where $\tau_{text}(text)$ is a CLIP~\cite{clip} text encoder that projects a sequence of tokenized texts into an embedding space. To jointly condition the image generation on hand parameters and the text, we replace the text encoder $\tau_{text}(text)$ with a novel Text+Hand encoder $\tau_{text+h}( text, hand)$ that jointly embeds the text and hand parameters into a common embedding space.

\subsection{\compA~Diffusion}

The \compA~diffusion model takes a text as input and generates the pose parameters $\theta = (\theta_b, \theta_{lh}, \theta_{rh})$ and shape parameters $\beta$ for the SMPL-H model by conditioning on the text. 

We define $x := (\theta, \beta)$ and model the forward diffusion process by iteratively adding Gaussian noise to $x$ for $T$ time steps:
\begin{equation}
    q\parens{x_t | x_{t-1}} = \mathcal{N}\parens{\sqrt{\alpha_t} x_{t-1}, (1 - \alpha_t) I},
\end{equation}
where $\alpha_t \in (0, 1)$ are constant hyper-parameters.

We model the text-conditioned SMPL-H generation distribution $p(x_0|c)$ as the reverse diffusion process of gradually denoising $x_T$. Following~\cite{MDM}, we learn the denoising by directly predicting $\hat{x}_0 = G(x_t, t, c)$ using a model $G$. We train the reverse diffusion using the training objective:
\begin{equation}
    \mL_1 = \mathbb{E}_{x_0 \sim q(x_0|c), t\sim [1, T]} || x_0 - G(x_t, t, c) ||^2_2.
\end{equation}

We get the conditional text embeddings $c$ by encoding the text using CLIP~\cite{clip}. We implement $G$ using a transformer encoder-only architecture similar to MDM~\cite{MDM}.

Given a text during inference, we conditionally sample $x = (\theta, \beta)$. We use the shape and pose parameters to obtain the joints $J_{lh}, J_{rh}$ and vertices $\mathcal{M}_{lh}, \mathcal{M}_{rh}$ for left and right hands using MANO-Hand model. We also choose camera parameters and project $J_{lh}, J_{rh}$ into an image space and obtain the corresponding image-space joint locations $J^{2D}_{lh}, J^{2D}_{rh}$. We use the joint rotations $\theta_{lh}, \theta_{rh}$, hand vertices $\mathcal{M}_{lh}, \mathcal{M}_{rh}$, and spatial joint locations $J^{2D}_{lh}, J^{2D}_{rh}$ to condition the image generation in the next stage. 

\subsection{\compB~Diffusion}

The \compB~diffusion model is built upon Stable Diffusion~\cite{Rombach_2022_CVPR} and conditions the image generation on hand parameters generated from the \compA~model and the text. Specifically, \compB~ uses a novel Text+Hand Encoder $\tau_{text+h}$ to first obtain joint embeddings for text and hand parameters. It then uses the joint hand and text embeddings to condition the image generation. We provide more details on this below.

\myheading{Text+Hand Encoder.} Given the provided text, along with the spatial joint locations $J^{2D}_h$, vertices $\mathcal{M}_h$, and joint rotations $\theta_h$ of the hand, our goal is to generate $D$-dimensional embeddings to encode both the text and hand parameters. Here $D$ denotes the CLIP~\cite{clip} token embedding dimension. To encode hand joint locations in the image space, we follow~\cite{chen2022pixseq, Yang_2023_CVPR} and introduce additional positional tokens. We quantize the image height and width uniformly into $N_{bins}$ bins. This allows us to approximate and tokenize any normalized spatial coordinate into one of $N_{\text{bins}}$ tokens. We then encode the text tokens and the hand joint spatial tokens into a $D$-dimensions using $f_{text+J^{2D}_h}$. Specifically, we construct $f_{text+J^{2D}_h}$ by introducing an additional $N_{\text{bins}} \times D$ embedding layer into an existing CLIP token embedder and finetune it during training. To encode hand vertices, we transform them to basis point set (BPS)~\cite{BPS_ICCV_19} representations and pass through $f_{\mathcal{M}_h}$, a Multi-Layer Perceptron (MLP)  consisting of fully-connected linear and ReLU layers. Similarly, we encode 6D hand joint rotations $\theta_h$ using an MLP $f_{\theta_h}$ consisting of fully-connected linear and ReLU layers. Finally, we concatenate embeddings from text, spatial hand joints, hand vertices, and hand joint rotations to produce joint text and hand embeddings.

\myheading{Diffusion.} We instantiate the \compB~ using Stable Diffusion~\cite{Rombach_2022_CVPR} and train using the following objective
\begin{equation}
    \mL_{2} = \mathbb{E}_{\mathcal{E}(x), \epsilon \sim \mathcal{N}(0, 1), t, y} || \epsilon - F\parens{z_t, t, \tau_{text+h}(y)}||^2_2. \label{eqn:img_diffusion}
\end{equation}
In the above equation, the condition $y = (\text{text}, J^{2D}_h, \mathcal{M}_h, \theta_h)$ denotes the combination of the text and the hand parameters, which include spatial joint locations, vertices, and joint rotations. The function $F$ is a denoising U-Net to predict the noise, $\tau_{text+h}$ is the trainable Text+Hand encoder. We refer the readers to ~\cite{Rombach_2022_CVPR} for more details regarding \Eref{eqn:img_diffusion}.

\setlength{\tabcolsep}{5pt}
\begin{table*}[ht]
    \centering
    \scalebox{1.0}{
        \begin{tabular}{lccccc}
        \toprule
        Method & FID $\downarrow$ & KID $\downarrow$ & FID-H $\downarrow$ & KID-H $\downarrow$ & Hand Conf. $\uparrow$ \\
        \midrule
        Stable Diffusion & $29.005$ & $9.63{\times}{10}^{-3}$ & $34.372$ & $4.63{\times}{10}^{-2}$ & $0.887$\\
        Stable Diffusion Fine-tuned  & $20.056$ & $7.91{\times}{10}^{-3}$ & $31.219$ & $3.09{\times}{10}^{-2}$ & $0.913$\\
        ControlNet & $18.694$ & $5.93{\times}{10}^{-3}$ & $28.091$ & $2.19{\times}{10}^{-2}$ & $0.969$ \\
        \midrule
        \modelname~w/o 2D hand joints & $16.839$ & $5.21{\times}{10}^{-3}$ & $29.902$ & $2.46{\times}{10}^{-2}$ & $0.953$ \\
        \modelname~w/o 3D joint rotation and vertices & $14.586$ & $4.14{\times}{10}^{-3}$ & $28.186$ & $2.21{\times}{10}^{-2}$ & $0.961$ \\
        \rowcolor{lightgray}
        \modelname~(proposed) & $\mathbf{13.918}$ & $\mathbf{4.07{\times}{10}^{-3}}$ & $\mathbf{27.550}$ & $\mathbf{2.11{\times}{10}^{-2}}$ & $\mathbf{0.978}$\\
        \bottomrule
        \end{tabular}
    }
    \caption{\textbf{Quantitative results.} We report the scores of current baselines and ablated versions of our method on multiple evaluation metrics. $\uparrow$ indicates higher values are better, $\downarrow$ indicates lower values are better.}
    \label{tab:quant_results}
    \vskip -.1in
\end{table*}

\myheading{Generating SMPL-H vs Skeletons in \compA~Diffusion.} We design the first component of \modelname~to generate pose and shape parameters of SMPL-H instead of keypoints or skeletons since SMPL-H encodes topological and geometric priors about humans and encodes richer information than skeletons. Also, SMPL-H parameters tend to be more robust to noise than skeletons; we can still get plausible poses even with noisy SMPL-H parameters, whereas noisy joint locations lead to implausible poses. Since we are generating \textit{parameters} (51 joint rotations and 10 shape parameters) of SMPL-H mesh, the \compA~component is computationally lighter compared to the second component, \compB~.

\begin{figure*}
    \centering
    \includegraphics[width=\linewidth]{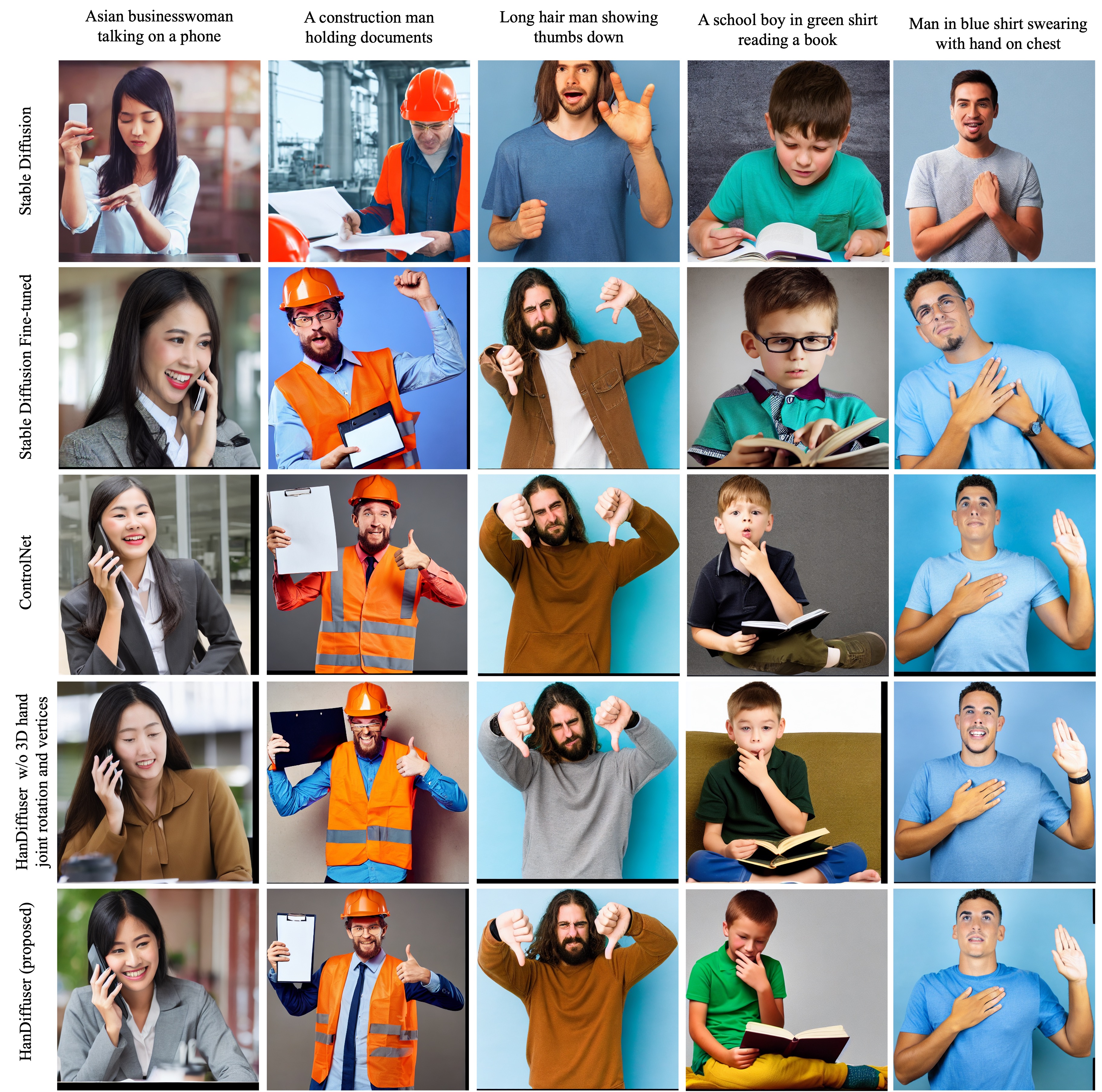}
    \caption{\textbf{Qualitative results.} We compare the quality of hands in images generated by different methods from the same text prompts. (Images are generated at 512x512 resolution)}
    \label{fig:qual_res}
    \vskip -.1in
\end{figure*}

\section{Experiments}

This section describes the datasets used to train \modelname~, implementation details, and evaluation metrics used. We also present qualitative and quantitative results, and user studies to show the efficacy of \modelname~ in generating images with high-quality hands.

\subsection{Datasets}
We train the two components of \modelname~ using our own curated datasets. We start with 930K paired text and images, then curate it to remove inappropriate and harmful content and validate the quality of the images through independent content creators. We randomly split the dataset to obtain 900K train and 30K test text-image pairs. We further preprocess the dataset to obtain SMPL-H parameters. Specifically, we use~\cite{ROMP} to obtain SMPL parameters for the body and~\cite{MANO_Hand} to obtain MANO parameters for hands. We reject estimated SMPL body and MANO hands that have low confidence scores. Finally, we curate two datasets using the estimated hand and body parameters. The first dataset consists of tuples of the form (text, SMPL-H). We keep such tuples only for images where we can reliably estimate SMPL-H parameters. This dataset has 450K tuples (text, SMPL-H) and is used to train the first component of the \modelname, \compA. The second dataset consists of triplets of the form (text, image, SMPL-H), and we keep all 930K triplets. We use this dataset to train the second component of \modelname, \compB. During training, we only conditioned the image generation on hand parameters when the SMPL-H parameters were reliably estimated.

\subsection{Implementation Details}
We train the \compA~diffusion to generate the SMPL-H pose $\theta$ and shape $\beta$ by conditioning on the text. We encode the text using a frozen CLIP-ViT-B/32 model~\cite{clip}. We train the \compA~model using a classifier-free guidance~\cite{ho2021classifierfree} by randomly setting $10\%$ of the text conditions to be empty. We train this model for 100 epochs on a single A100 GPU using a batch size of 64. We use 1000 steps and a guidance scale of $s=2.5$ during the inference. We fine-tune \compB~ starting from the Stable Diffusion v1.4 checkpoint. To implement the Text+Hand encoder $\tau_{text+h}$, we start with the CLIP ViT-L/14 model and introduce additional $N_{bins}=1000$ positional tokens for spatial hand joints. We choose simple three-layer MLPs $f_{\mathcal{M}_h}$ and $f_{\theta_h}$ to encode hand vertices and joint rotations, respectively. We fine-tune \compB, including the Text+Hand encoder $\tau_{text+h}$, for 20 epochs on eight A100 GPUs using a batch size of 8 and AdamW optimizer with a constant learning rate of $10^{-4}$. We perform inference with 50 PLMS~\cite{plms} steps using a classifier-free guidance~\cite{ho2021classifierfree} of 4.0.

\myheading{\modelname~Inference.} Given a text input, we first sample SMPL-H parameters using our trained \compA~model. We then extract the MANO hand parameters from SMPL-H and choose camera parameters randomly with some constraints to make hands somewhat visible in the image and obtain spatial hand joint joints. Finally, we use these spatial hand joints, MANO parameters, and the text to conditionally sample an image from our trained \compB~model.

\subsection{Evaluation Metrics}
We access the quality of generated images from \modelname~ using the  Frechet Inception Distance (FID) and Kernel Inception Distance (KID)~\cite{FID, clean_fid}. Since FID and KID measures the overall quality of the image, we also compute FID-H and KID-H to measure the quality of images only in the hand regions. We perform this by first extracting crops using hand bounding boxes and then computing FID and KID using such hand crops. We also measure the quality of hands using average hand detection confidence scores. Specifically, we run an off-the-shelf hand detector~\cite{zhang_arxiv_2020} on generated images and compute the confidence scores for detection. Higher confidence scores mean that the hand detector is more confident of a region being a hand, indicating higher-quality hand generations.

\subsection{Quantitative Results and Ablation Studies}
We compare the proposed \modelname~with three different methods and report these results in \Tref{tab:quant_results}. First, we use an off-the-shelf Stable Diffusion~\cite{Rombach_2022_CVPR} model pre-trained on the LAION-5B~\cite{laion5b} dataset. The LAION-5B dataset is a general-purpose text and image pairs dataset and does not necessarily focus on humans. Therefore, a Stable Diffusion model trained on the LAION-5B dataset does not perform well on images solely focused on humans. Second, we fine-tuned Stable Diffusion on our dataset and observed that it generated better images than the pre-trained model. While this generates better performance than the pre-trained model, the performance is still low compared to the proposed \modelname. Third, we experiment with ControlNet~\cite{controlnet}, a popular latent diffusion model that uses spatial control images to condition the image generation process. We train a ControlNet architecture on our dataset using hand-pose skeleton images as controls. However, unlike \modelname, which generates images directly from text input, ControlNet requires an additional hand pose skeleton image as input during inference. To address this, we directly use the \textit{ground-truth} skeleton images from the test data as control images. Despite employing these \textit{ground-truth} control images, ControlNet does not perform as well as \modelname. 

It is important to note that the reported FID-H, KID-H, and hand confidence scores for Stable Diffusion and Stable Diffusion Fine-tuned in \Tref{tab:quant_results} are optimistic performance measures. To evaluate the performance of these two methods, we first ran a hand detector~\cite{zhang_arxiv_2020} to obtain hand crops. However, this approach is biased towards rejecting bad hand generations since the hand detector cannot localize low-quality hand generations, leaving out unrealistic-looking generated hands from evaluation. On the contrary, the corresponding metrics for ControlNet and \modelname~in \Tref{tab:quant_results} are the true performance measures since both the methods generate images conditioned on hands, allowing us to crop every generated hand precisely.

We also study the benefits of different hand representations that are used to condition the hand generation in \modelname. First, we evaluate \modelname~by omitting the spatial hand joint locations $J^{2D}_h$ in hand embeddings. Second, we evaluate \modelname~by omitting the hand joint rotations $\theta_h$ and hand vertices $\mathcal{M}_h$ in hand embeddings. We report these results in the fourth and fifth row of \Tref{tab:quant_results}. These results show that all three hand representations help in generating quality hands. 

\begin{figure}[t]
\centering 
    \centering
    \includegraphics[width=\columnwidth]{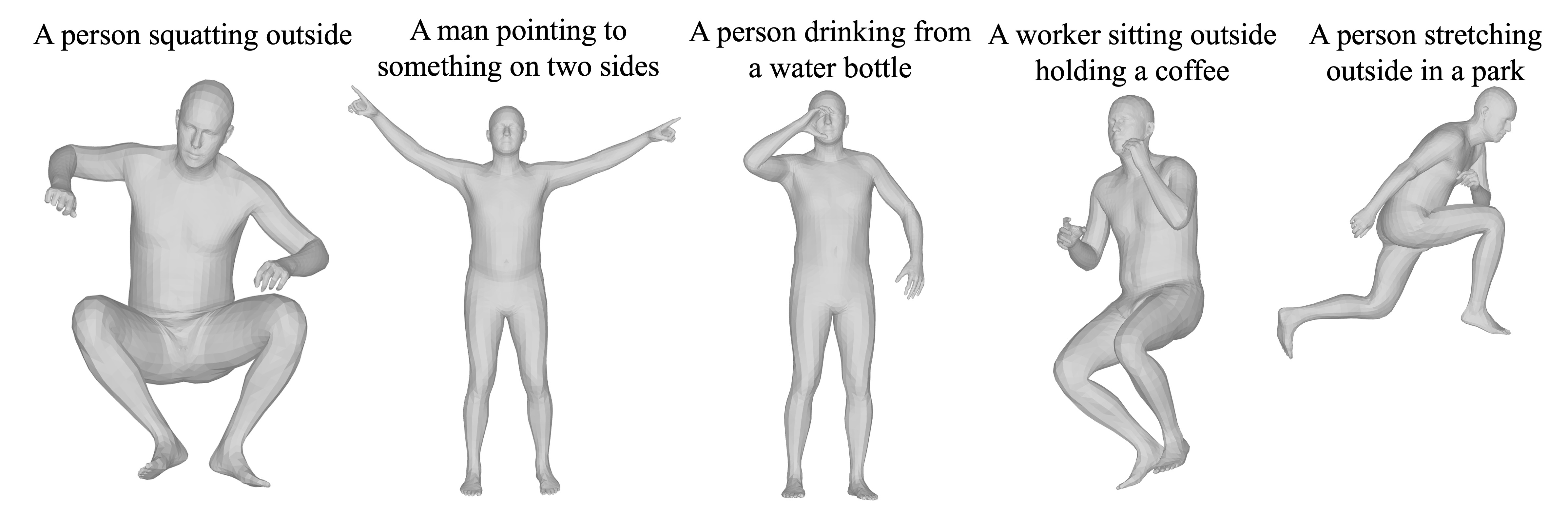}

    \caption{\textbf{Illustrative SMPL-H results},  generated from our \compA~model.} \label{fig:compA_results}
    \vskip -.1in

\end{figure}

\subsection{Qualitative Results and Failure Cases}
We report some good qualitative results in \Fref{fig:qual_res}. We compare results from Stable Diffusion, ControlNet~\cite{controlnet}, the proposed \modelname~without hand joint rotations and vertices embeddings, and \modelname. Stable Diffusion does not generate realistic hands even after fine-tuning on human-centric datasets. It generates an incorrect number of hand fingers, poor hand-object interactions, implausible finger orientations, and hand shapes. ControlNet generates better-looking results but requires hand skeleton control images as additional input. We can see that \modelname~generates hands with plausible hand poses by conditioning the image generation on spatial hand joint locations. Further conditioning on hand joint rotations and vertices enables \modelname~to generate high-quality, detailed hands with plausible orientations and shapes. 

\begin{figure}[t]
\centering 
    \centering
    \includegraphics[width=\columnwidth]{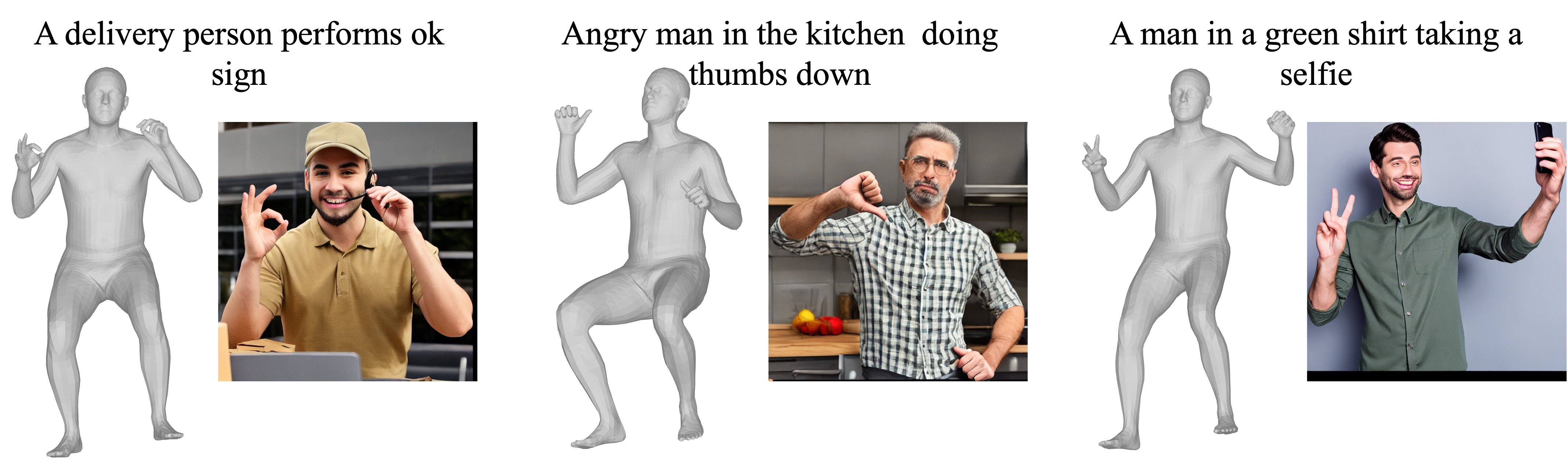}

    \caption{\textbf{Generating images from text via SMPL-H.} The intermediate SMPL-H representations are essential in generating realistic hand appearances.} \label{fig:text_smpl_img}
    \vskip -.1in
\end{figure}

\begin{figure*}
    \centering 
    \includegraphics[width=\textwidth]{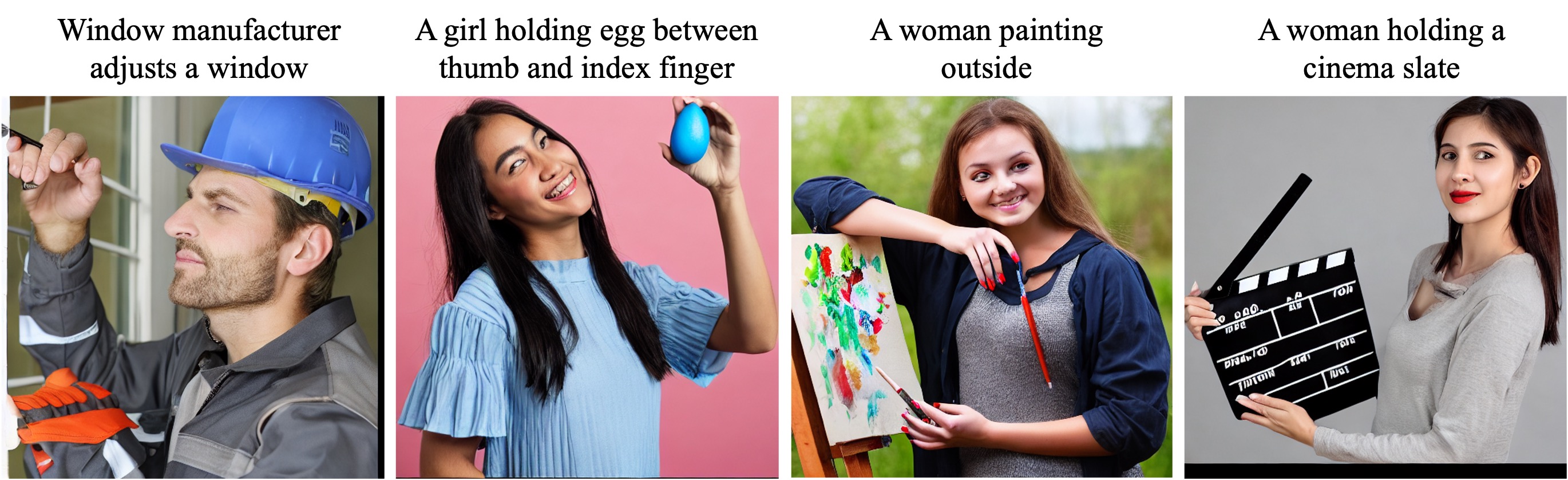}

    \caption{\textbf{\modelname~failure cases.} We note some failure cases when the given action may be unclear (\textit{e.g.}, ``adjusts'' in the first image from left), the model does not exactly follow the hand pose description (second image from left), the model does not fully realize the finger dexterity when handling small and thin objects (third image from left), and model does not strictly obey the intended affordance of the object (fourth image from left).}
    \label{fig:failure_cases}
    \vskip -.1in
\end{figure*}

\Fref{fig:compA_results} shows a few SMPL-H results generated from text inputs using our \compA. While we only use hand parameters from these SMPL-H outputs, our \compA~ can be directly used in other applications that require generating SMPL-H models from text inputs. We also show how \compB~maps these SMPL-H results to generated images in \Fref{fig:text_smpl_img}. \Fref{fig:failure_cases} shows some failure cases of \modelname.

\subsection{User Studies}
We evaluate the quality of both our generated images and the intermediate outputs of our approach through two different user studies. We evaluate the generated images in two aspects. The first is \textit{(A) plausibility}, which considers how natural the hands look, for example, in terms of hand shapes, finger orientations, number of fingers, hands, and how clearly the hands are in focus in the image. The second is \textit{ (B) relevance}, which considers how natural the hand poses or gestures appear given the prompt, for example, holding objects or gesticulating conventionally (unless otherwise specified in the prompt).

We evaluate the intermediate SMPL-H outputs for generating the images in three aspects: \textit{(A) plausibility} of the pose, \textit{(B) relevance} to prompt, and \textit{(C) consistency} with the generated image.

\begin{figure}[t]
    \centering
    \begin{subfigure}[b]{\columnwidth}
        \centering
        \includegraphics[width=\textwidth]{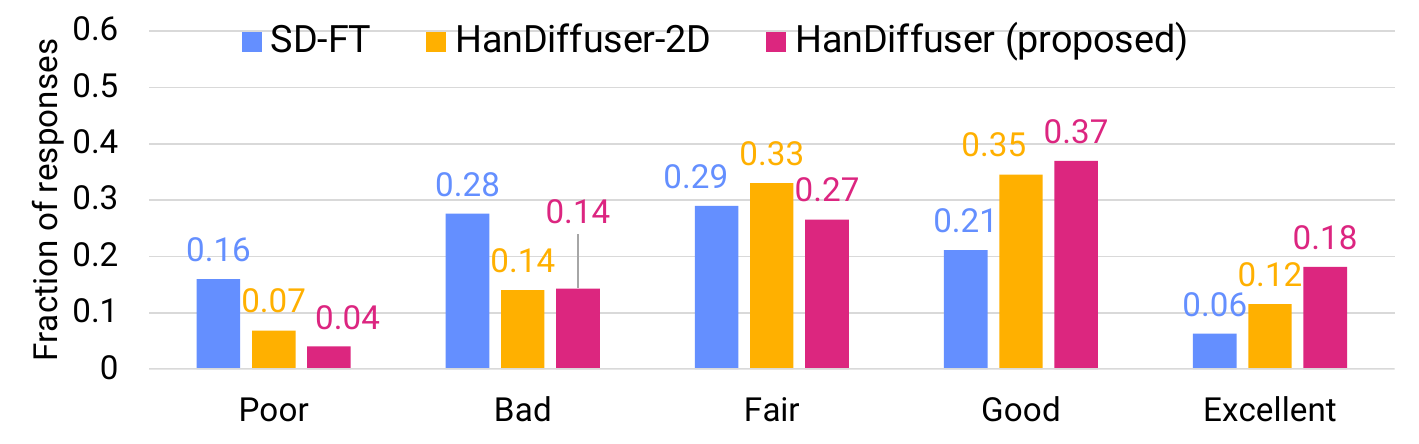}
        \caption{Results on the plausibility of generated images}
        \label{fig:user_study_results_plausibility_images}
    \end{subfigure}
    \begin{subfigure}[b]{\columnwidth}
        \centering
        \includegraphics[width=\textwidth]{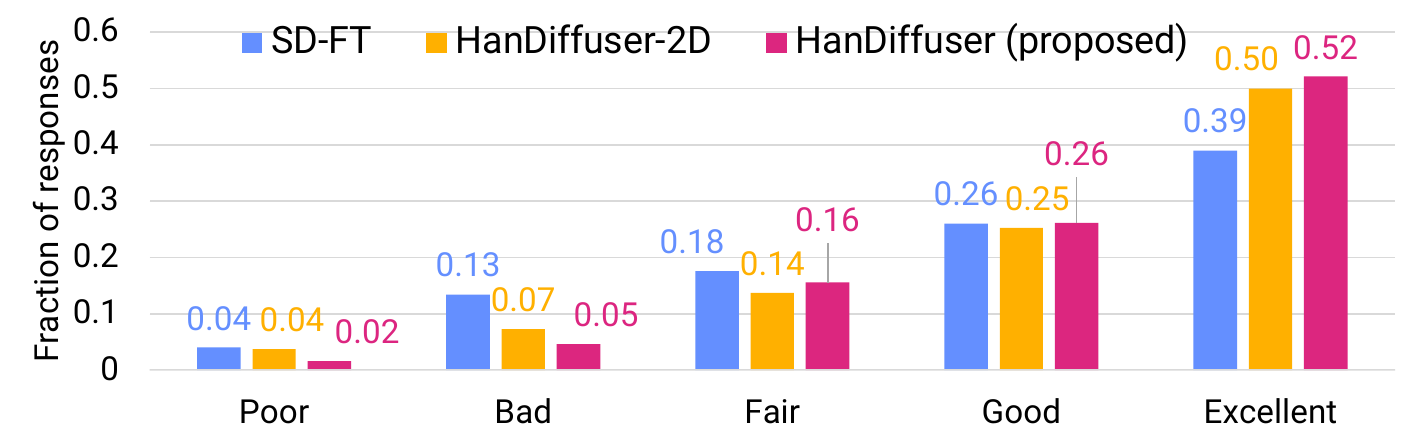}
        \caption{Results on the relevance of generated images to given prompts}
        \label{fig:user_study_results_relevance_images}
    \end{subfigure}
    \caption{\textbf{User study results for generated images.} We report the mean fraction of responses for each point on the Likert scale.}
    \label{fig:user_study_results_images}
    \vskip -.1in
\end{figure}

\myheading{Setup.}
We compare three methods in the user study to evaluate the generated images: fine-tuned Stable Diffusion (SD-FT), \modelname~trained with only 2D hand joints (\modelname-2D), and \modelname~trained with all its model components. We show participants 20 sets of images. Each set consists of a unique prompt randomly selected from the test partition of LAION-5B~\cite{laion5b} and the images generated by the three methods given that prompt. We arrange the three images within each set in a random order not known to the participants. For each image in each set, we ask them to respond to two questions: \textit{``How is the visual quality of the hands?''} (image plausibility) and \textit{``How well do the hands follow the prompt?''} (image relevance). We collect responses to the two questions on a 5-point Likert scale, consisting of the following choices: \textit{``Poor (e.g., too many or severe mistakes)''}, \textit{``Bad (e.g., some aspects reasonable but still many or severe mistakes)''}, \textit{``Fair (e.g., some aspects are plausible but some mistakes visible)''}, \textit{``Good (e.g., most aspects are plausible but a few mistakes visible)''}, and \textit{``Excellent (e.g., everything looks good, no visible mistakes)''}.
Note that we perform the user study on methods that only require text prompts to generate image outputs at test time. Therefore, we exclude methods such as ControlNet~\cite{controlnet}, which additionally requires pose information to generate similar images. Moreover, the overall performance of ControlNet is also at the same level as \modelname-2D (\Tref{tab:quant_results} rows 3 and 4) even if we manually provide the ground-truth poses, leading to no meaningful differences between their responses in a pilot study.

\begin{table}[t]
    \centering
    \resizebox{\columnwidth}{!}{
        \begin{tabular}{lccc}
        \toprule
        Image Aspect & SD-FT & \modelname-2D & \modelname~(proposed) \\
        \midrule
        Plausibility $\uparrow$ & $2.74 \pm 0.08$ & $3.30 \pm 0.11$ & $\mathbf{3.51 \pm 0.11}$ \\
        Relevance $\uparrow$ & $3.83 \pm 0.12$ & $4.11 \pm 0.17$ & $\mathbf{4.23 \pm 0.18}$ \\
        \bottomrule
        \end{tabular}
    }
    \caption{\textbf{User study for generated images score summary.} We compute the mean and standard deviation of the Likert-scale scores of the three evaluated methods across all the 700 responses of 35 participants. Higher scores are better.  \label{tab:user_study_scores_images}}
\end{table}

For the user study to evaluate the intermediate SMPL-H outputs, we show participants 9 random triplets of (prompt, SMPL-H poses, generated image), where the prompts are randomly selected from the test partition of LAION-5B~\cite{laion5b} and the SMPL-H poses and generated images come from our approach. For each triplet, we ask the participants to respond to three questions: \textit{``How plausible is the pose?''} (SMPL-H plausibility), \textit{``How relevant are the hands in the pose given the prompt?''} (SMPL-H relevance), and \textit{``How consistent are the hands in the pose with the hands in the image?''} (SMPL-H consistency). We collect responses to the three questions on the same 5-point Likert scale as above. ask them to evaluate the poses on 5-point Likert scales for each of the three aspects of plausibility, relevance, and consistency. To evaluate consistency, we additionally ask participants to focus primarily on the hand configurations and gestures described or implied in the text. We ask them to ignore distractors, such as the quality of any facial expressions (or lack thereof), any component of the 3D pose that is not visible in the image, and the differences in orientations of the body between the pose and the image.

\begin{figure}[t]
    \centering
    \includegraphics[width=\columnwidth]{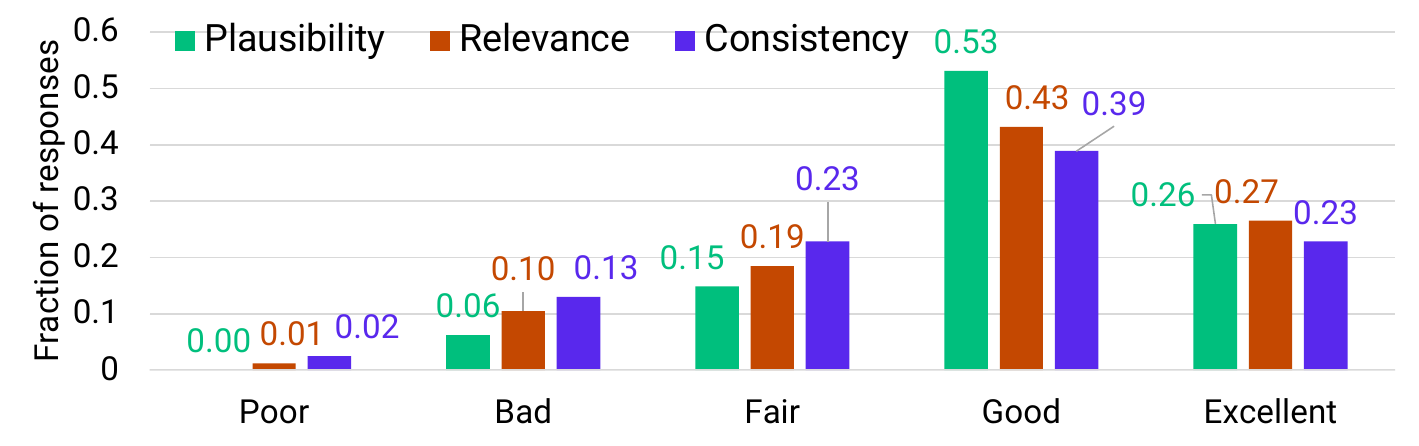}
    \vskip -.1in
    \caption{\textbf{User study results for SMPL-H poses.} We report the mean fraction of responses for each point on the Likert scale.}
    \label{fig:user_study_results_smplh}
    \vskip -.2in
\end{figure}

\myheading{Results.}
Our user study to evaluate the generated images was completed by 35 participants, resulting in a total of 700 responses over the 20 image sets. We did not observe any notable response differences across genders and age groups. We report the distribution of scores for the two aspects across all the responses in \Fref{fig:user_study_results_images}. We also summarize the mean and the standard deviation of the scores of the two aspects for each of the three methods in \Tref{tab:user_study_scores_images}. To compute these values, we assign numbers $1$ through $5$ to the response choices \textit{Poor} through \textit{Excellent}. Consequently, higher scores indicate better performance.
\modelname~outperforms the other methods in both aspects for generated images. Looking at the distribution of image plausibility scores (\Fref{fig:user_study_results_plausibility_images}), we observe the mode of SD-FT on \textit{``Fair''}, while the modes of both \modelname~versions are a point higher, on \textit{``Good''}. Overall, $55\%$ of \modelname~scores are \textit{``Good''} or better, compared to $47\%$ of \modelname-2D scores and $27\%$ of SD-FT scores. Looking at the distribution of image relevance scores (\Fref{fig:user_study_results_relevance_images}), we observe the modes of all the three methods on \textit{``Excellent''}, indicating their efficacy in generating hand appearances aligned with text prompts. Among the three methods, we note a relatively higher distribution of good responses for HandDiffuser variants. Specifically, $78\%$ of \modelname~scores are \textit{``Good''} or better, compared to $75\%$ of \modelname-2D scores and $65\%$ of SD-FT scores. Looking at the mean scores across all the 700 responses (\Tref{tab:user_study_scores_images}), we note marked improvements for \modelname. Its image plausibility scores are $0.77$ points (or $15\%$ on the 5-point scale) higher than SD-FT and $0.21$ points (or $4\%$) higher than \modelname-2D. Correspondingly, its image relevance scores are $0.40$ points (or $8\%$) higher than SD-FT and $0.12$ points (or $2\%$) higher than \modelname-2D.

Our user study to evaluate the intermediate SMPL-H poses was completed by 18 participants, resulting in a total of 171 responses over the 9 triplets. We did not observe any notable response differences across genders and age groups. We report the distribution of scores for the two aspects across all the responses in \Fref{fig:user_study_results_smplh}. We also summarize the mean and the standard deviation of the scores of the two aspects for each of the three methods in \Tref{tab:user_study_scores_smplh}. To compute these values, we assign numbers $1$ through $5$ to the response choices \textit{Poor} through \textit{Excellent}. Consequently, higher scores indicate better performance.

\begin{table}[t]
    \centering
    \begin{tabular}{ccc}
    \toprule
    Plausibility $\uparrow$ & Relevance $\uparrow$ & Consistency $\uparrow$ \\
    \midrule
    $4.00 \pm 0.81$ & $3.83 \pm 0.98$ & $3.67 \pm 1.04$ \\
    \bottomrule
    \end{tabular}
    \vskip -.05in
    \caption{\textbf{User study for SMPL-H score summary of \modelname.} We compute the mean and standard deviation of the Likert-scale scores for the three evaluation aspects across all 162 responses of 18 participants. Higher scores are better. \label{tab:user_study_scores_smplh}}
    \vskip -.2in

\end{table}

\section{Conclusions and Limitations}
We have presented \modelname, an end-to-end model to generate images with realistic hand appearances from text prompts. Our model explicitly learns hand embeddings based on hand shapes, poses and finger-level articulations, and combines them with text embeddings to generate images with high-quality hands. We demonstrate the state-of-the-art performance of our method on the benchmark T2I dataset both quantitatively, through multiple evaluation metrics, and qualitatively, through a user study.

In the future, we plan to extend our model to more unexplored territories of hand generation. These include images consisting of multiple people, complex hand-object interactions, prompts describing highly specialized hand activities (\textit{e.g.}, origami), the same person handling multiple objects simultaneously, hand-hand interactions of two or more people, and non-anthropomorphic hands (\textit{e.g.}, a dog a using a computer). A concurrent future direction is to make the hand generation pipeline style- and shape-aware, such that it can consistently generate the same hands when asked to generate the same person in different images.

{\small    \myheading{Acknowledgements.} This project was partially supported by US National Science Foundation Award NSDF DUE-2055406.}

{
    \small
    \bibliographystyle{ieeenat_fullname}
    \bibliography{main, sn_pubs}
}

\section{Additional Results}

\begin{figure*}[t]
    \centering
    \includegraphics[width=\linewidth]{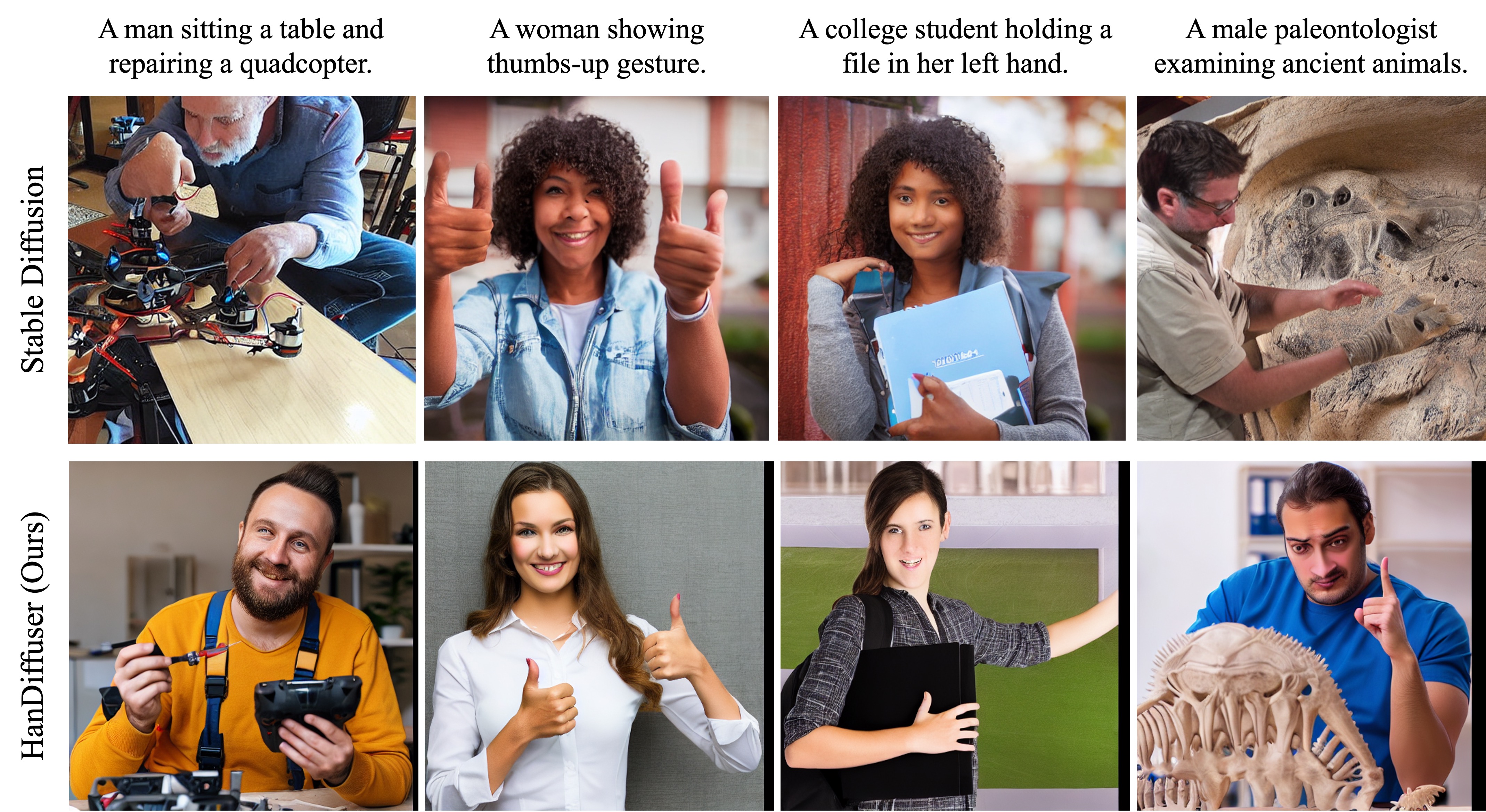}
    \vskip 0.5in
    \includegraphics[width=\linewidth]{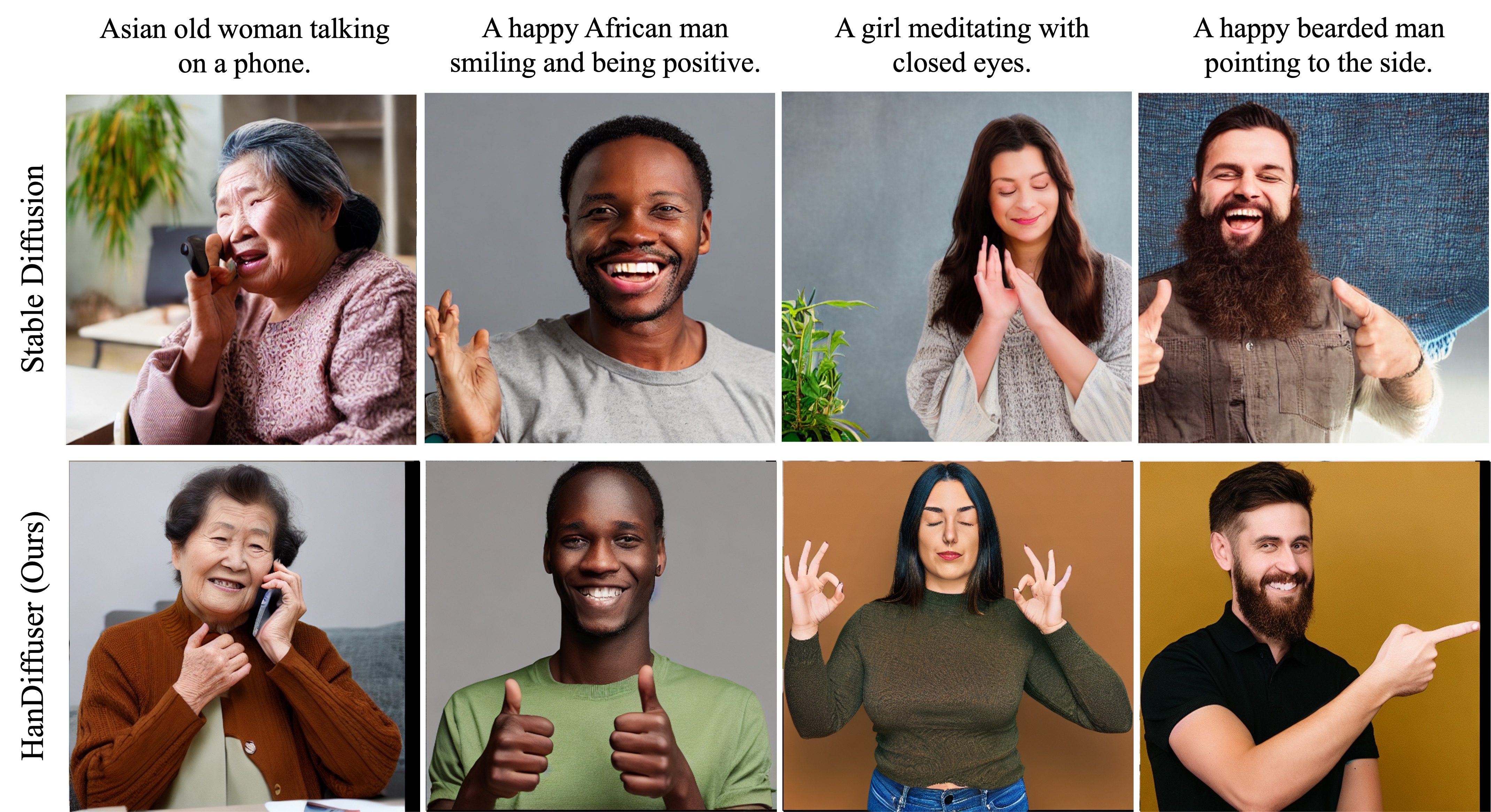}

    \caption{\textbf{Qualitative Comparison.} We compare the quality of hands in images generated from Stable Diffusion and the proposed \modelname.}
    \label{fig:qual_res_1}
\end{figure*}

\begin{figure*}[t]
    \centering
    \includegraphics[width=\linewidth]{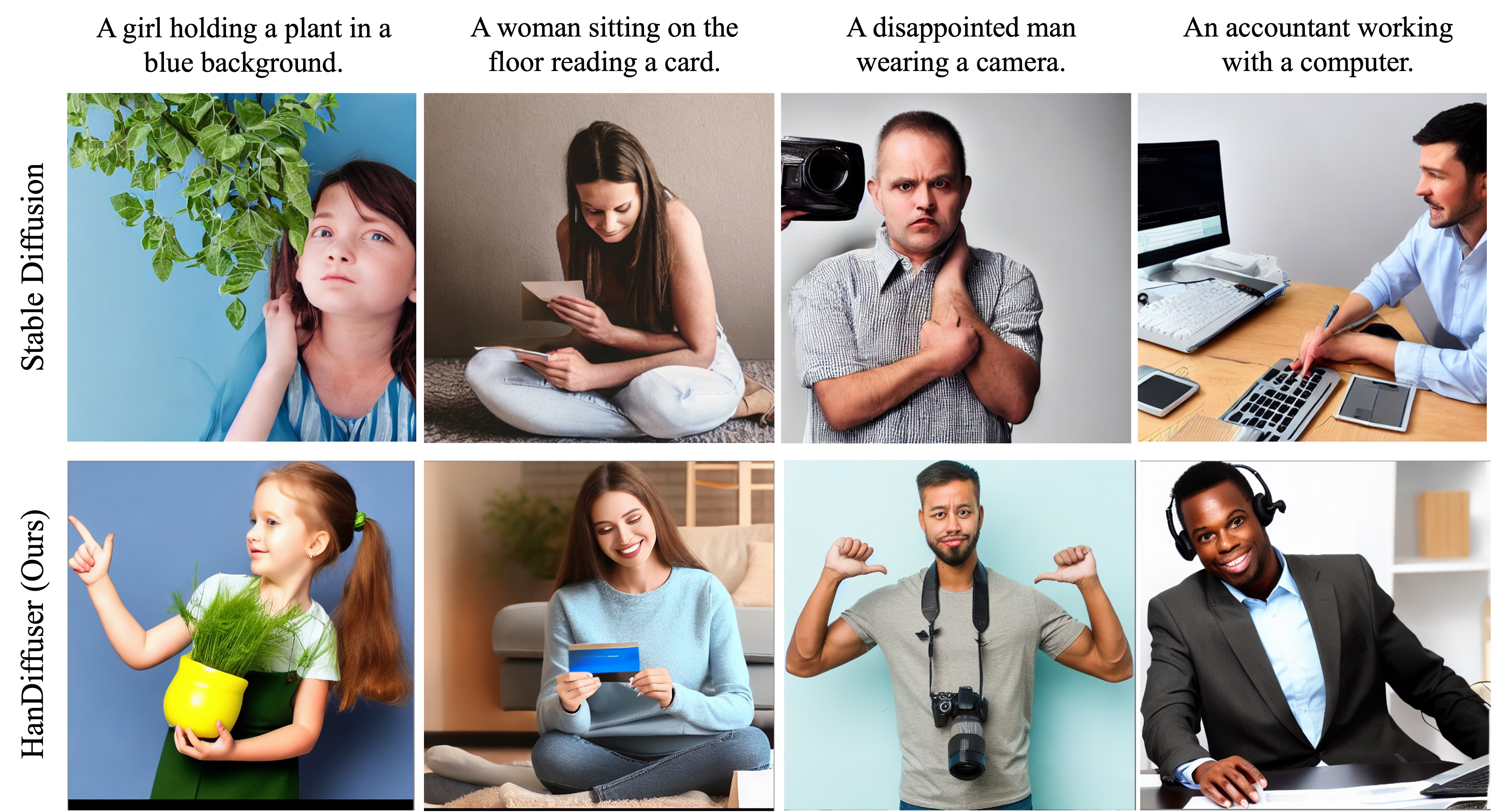}
    \vskip 0.5in
    \includegraphics[width=\linewidth]{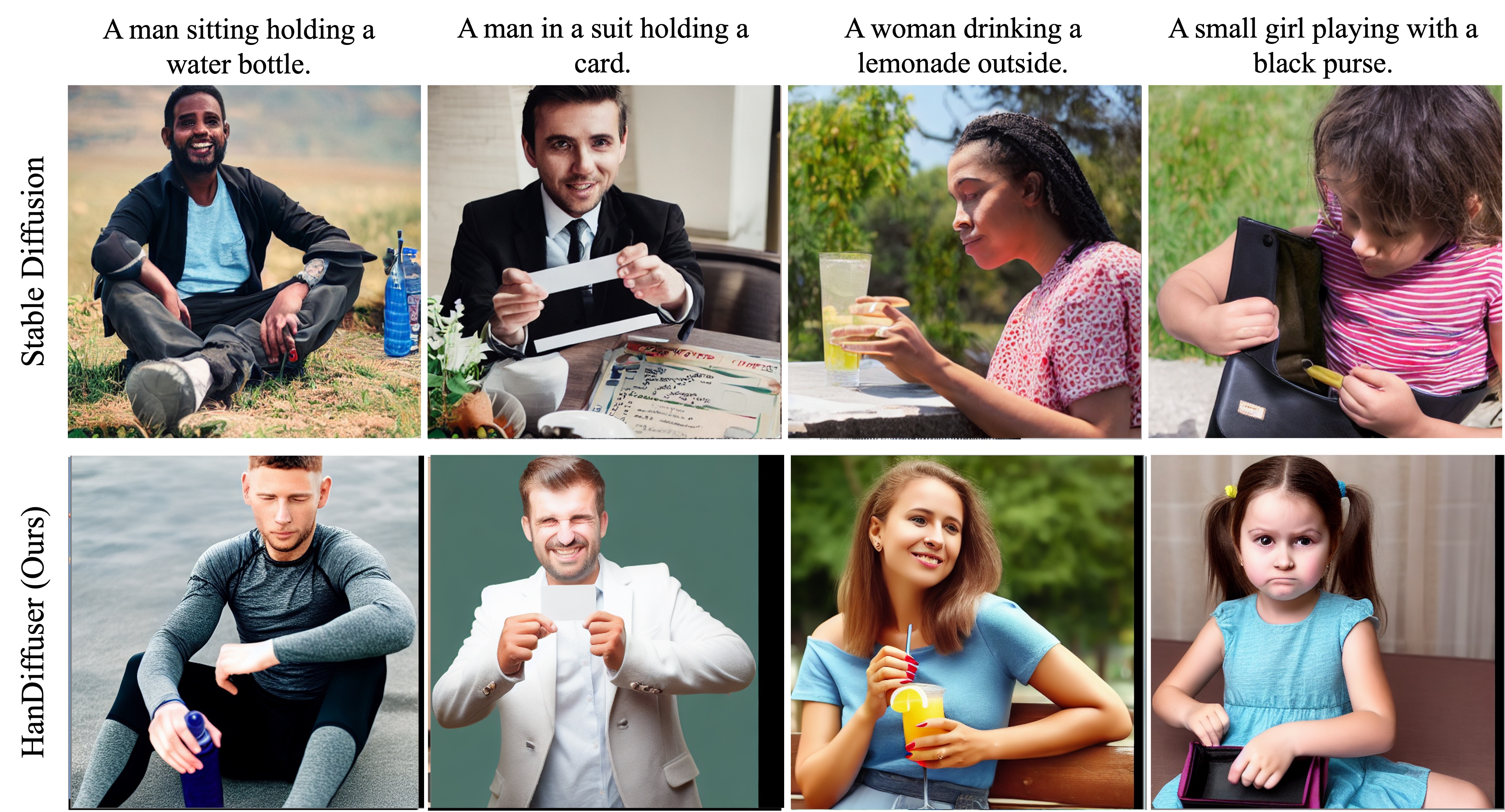}

    \caption{\textbf{Qualitative Comparison.} We compare the quality of hands in images generated from Stable Diffusion and the proposed \modelname.}
    \label{fig:qual_res_2}
\end{figure*}

We show additional qualitative comparisons with the Stable Diffusion~\cite{Rombach_2022_CVPR} baseline and the proposed \modelname~in \Fref{fig:qual_res_1} and \Fref{fig:qual_res_2}. 

\begin{figure*}[t]
    \centering
    \includegraphics[width=\linewidth]{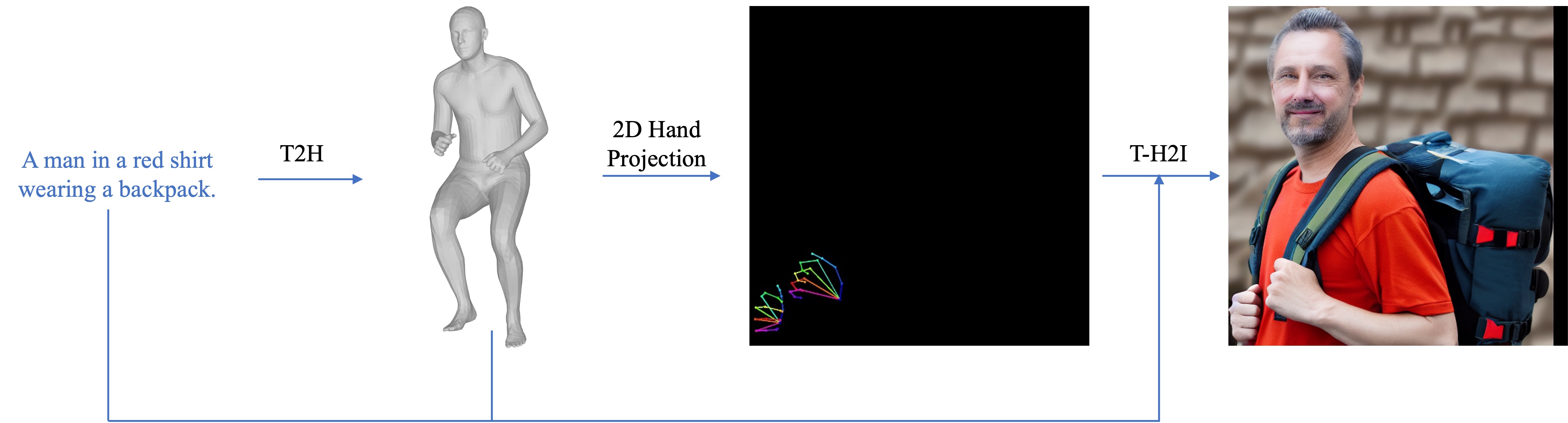}
    \vskip 0.5in
    \includegraphics[width=\linewidth]{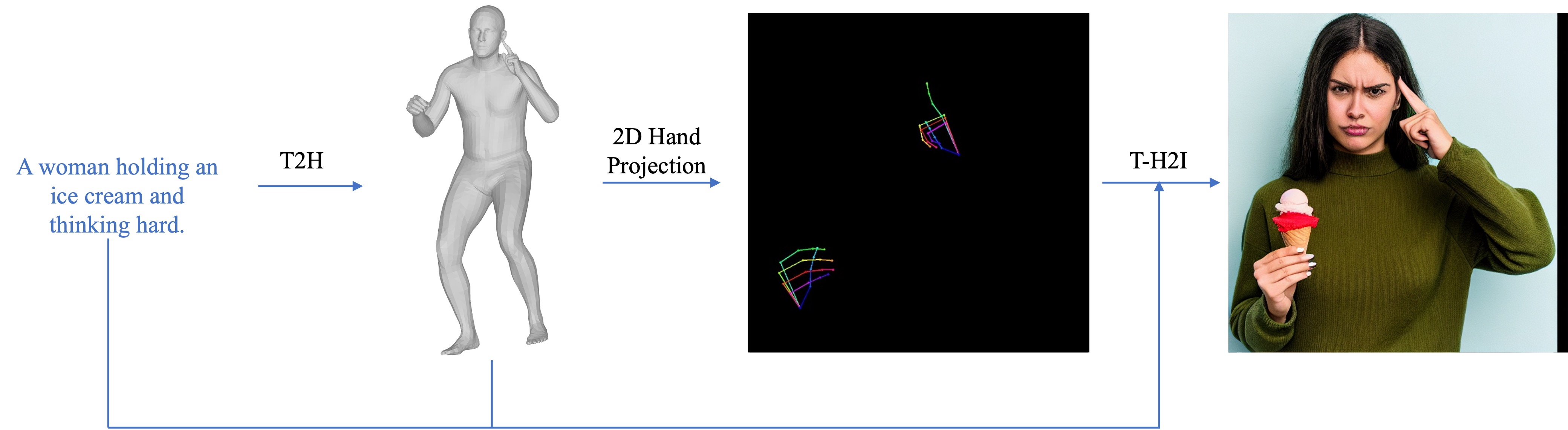}
    \vskip 0.5in

    \includegraphics[width=\linewidth]{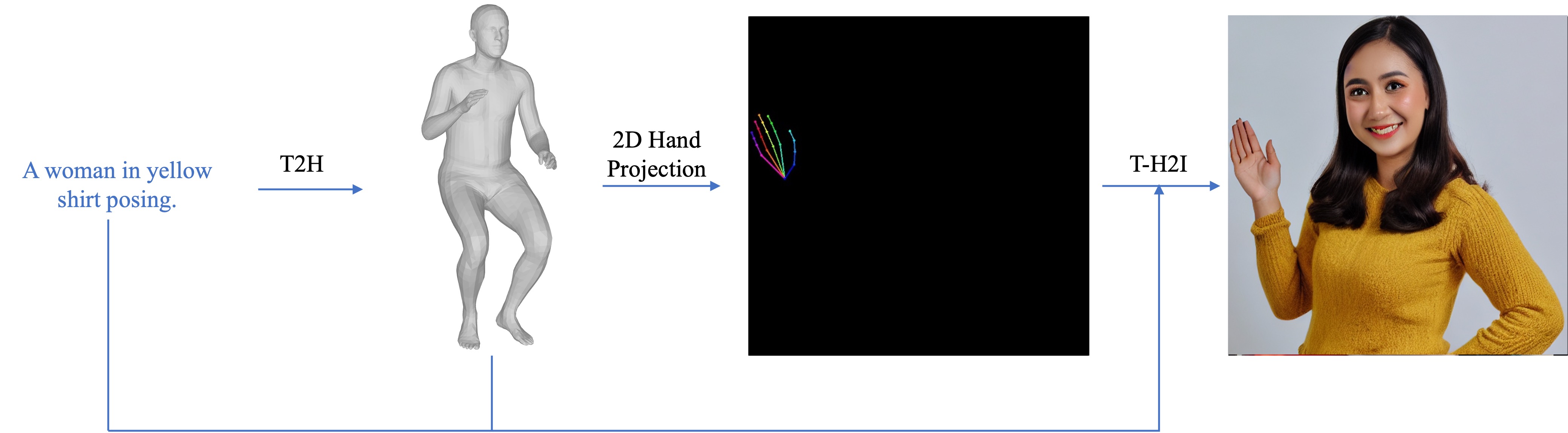}
    \caption{\textbf{\modelname~Qualitative Results.} Given a text input, \compA~(T2H) generates SMPL-H~\cite{SMPL_2015, MANO_Hand} parameters. We extract the MANO-Hand from SMPL-H and use some camera parameters to obtain 2D hand poses. The text, MANO-Hand and 2D hand poses are used to generate the final image using \compB~.}
    \label{fig:qual_res_3}
\end{figure*}

\begin{figure*}[t]
    \centering
    \includegraphics[width=\linewidth]{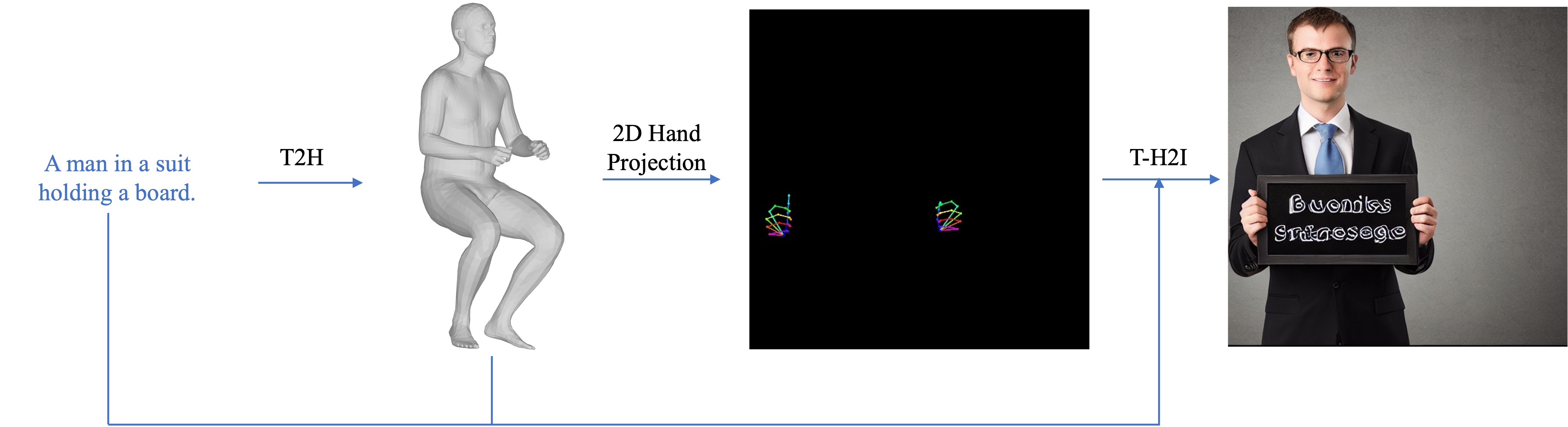}
    \vskip 0.5in
    \includegraphics[width=\linewidth]{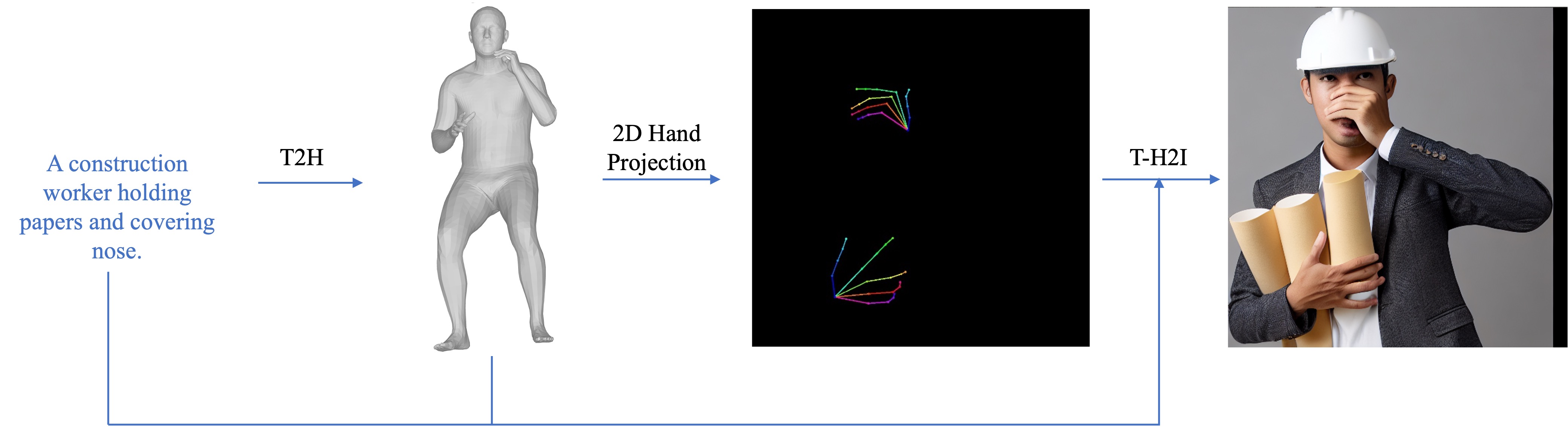}
    \vskip 0.5in
    \includegraphics[width=\linewidth]{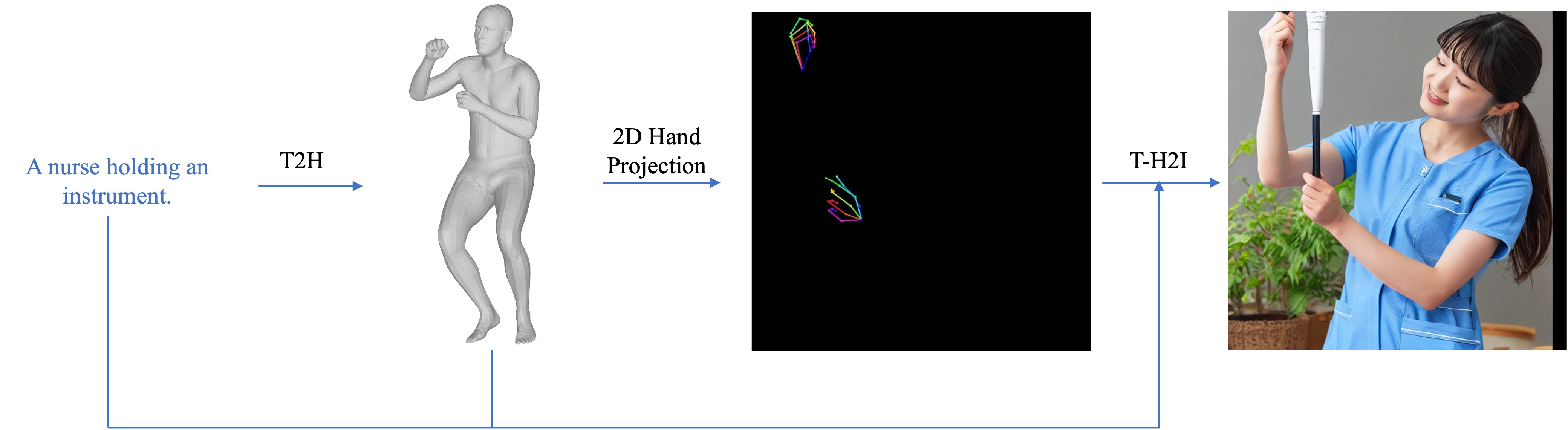}

    \caption{\textbf{\modelname~Qualitative Results.} Given a text input, \compA~(T2H) generates SMPL-H~\cite{SMPL_2015, MANO_Hand} parameters. We extract the MANO-Hand from SMPL-H and use some camera parameters to obtain 2D hand poses. The text, MANO-Hand and 2D hand poses are used to generate the final image using \compB~.}    \label{fig:qual_res_4}
\end{figure*}

We also show additional qualitative results and intermediate outputs from the two components of \modelname~ in \Fref{fig:qual_res_3} and \Fref{fig:qual_res_4}.

\end{document}